\PassOptionsToPackage{unicode}{hyperref}
\PassOptionsToPackage{hyphens}{url}
\PassOptionsToPackage{dvipsnames,svgnames,x11names}{xcolor}
\documentclass[
  11pt,
  a4paper,
]{article}
\usepackage{amsmath,amssymb}
\usepackage{iftex}
\ifPDFTeX
  \usepackage[T1]{fontenc}
  \usepackage[utf8]{inputenc}
  \usepackage{textcomp} 
\else 
  \usepackage{unicode-math} 
  \defaultfontfeatures{Scale=MatchLowercase}
  \defaultfontfeatures[\rmfamily]{Ligatures=TeX,Scale=1}
\fi
\usepackage{lmodern}
\ifPDFTeX\else
\fi
\IfFileExists{upquote.sty}{\usepackage{upquote}}{}
\IfFileExists{microtype.sty}{
  \usepackage[]{microtype}
  \UseMicrotypeSet[protrusion]{basicmath} 
}{}
\makeatletter
\@ifundefined{KOMAClassName}{
  \IfFileExists{parskip.sty}{%
    \usepackage{parskip}
  }{
    \setlength{\parindent}{0pt}
    \setlength{\parskip}{6pt plus 2pt minus 1pt}}
}{
  \KOMAoptions{parskip=half}}
\makeatother
\usepackage{xcolor}
\usepackage[margin=1in]{geometry}
\usepackage{graphicx}
\makeatletter
\def\maxwidth{\ifdim\Gin@nat@width>\linewidth\linewidth\else\Gin@nat@width\fi}
\def\maxheight{\ifdim\Gin@nat@height>\textheight\textheight\else\Gin@nat@height\fi}
\makeatother
\setkeys{Gin}{width=\maxwidth,height=\maxheight,keepaspectratio}
\makeatletter
\def\fps@figure{htbp}
\makeatother
\providecommand{\tightlist}{%
  \setlength{\itemsep}{0pt}\setlength{\parskip}{0pt}}
\ifLuaTeX
  \usepackage{selnolig}  
\fi
\usepackage{bookmark}
\IfFileExists{xurl.sty}{\usepackage{xurl}}{} 
\hypersetup{
  pdftitle={Do Not Trust the Benchmark: Limitations of General LLM Rankings and a Case for Task-Specific Evaluation},
  pdfauthor={Danial Amin},
  colorlinks=true,
  linkcolor={black},
  filecolor={Maroon},
  citecolor={Blue},
  urlcolor={blue},
  pdfcreator={LaTeX via pandoc}}

\title{Do Not Trust the Benchmark: Limitations of General LLM Rankings
and a Case for Task-Specific Evaluation}
\author{Danial Amin}
\date{19 September 2026}

\begin{document}
\maketitle

\begin{abstract}

Benchmark scores inform the development, marketing, and selection of
large language models (LLMs). Yet an overall score is interpretable only
in relation to the system tested, the questions included, and the
conditions of evaluation. This perspective examines five connected
limitations of general LLM rankings: differences between evaluated and
publicly available systems; commercial incentives and dependencies in
external evaluation; benchmark saturation, defective tests, and data
contamination; models exploiting scoring procedures; and the limited
relevance of general scores to users' tasks. Documented cases illustrate
why these problems require different responses. I argue for evaluation
procedures that disclose the tested configuration, validate questions
and successful task completion, report performance alongside cost and
execution time, and make the scope of generalization explicit. I then
discuss \textbf{Isotanta}, a crowdsourced benchmarking platform, as a
practical example of contributed questions and repeated evaluation. A
larger question pool may improve task coverage, while repeated sampling
can improve the stability of estimates on that pool; neither guarantees
validity or personalization. The paper distinguishes the platform's
current shared ranking from proposed task-specific and user-provided
evaluations. The central argument is that general rankings can inform
model selection but do not replace evidence about performance on the
intended work.

\end{abstract}

\noindent\textbf{Keywords:} large language models; benchmarking; evaluation
validity; crowdsourcing; task-specific evaluation

\subsection{1. Introduction}\label{introduction}

Large language models (LLMs) have become part of everyday and
professional life. People use them to plan trips, edit text written in a
second language, develop software, analyze information, and assist with
research. In knowledge-intensive work, models are increasingly
integrated into activities that previously depended on an individual's
ability to write, code, interpret evidence, or solve problems without AI
assistance. This is not necessarily undesirable: a useful tool can
improve how work is done. It does, however, make the ability to assess
the tool more important. \textbf{As dependence on LLMs increases, users
need reliable evidence about what a model can do for the tasks they
intend to perform.}

Consider a researcher selecting a model to analyze interview transcripts
or an engineer selecting one to work on an existing codebase. They can
consult a public leaderboard, but a high position does not establish
that the model follows their instructions, handles their inputs
accurately, or produces an improvement worth its cost. Benchmarks are
intended to support comparisons by giving models common tasks and
scoring their responses. Their scores, however, depend on the model
configuration, the question pool, the procedure used to judge
correctness, and the conditions under which the evaluation was
conducted.

This paper develops a \emph{perspective}, not a systematic review or a
new empirical comparison of LLMs. It draws on selected, publicly
documented cases to distinguish five problems: commercial incentives and
uncertainty about which system was evaluated; limitations of independent
access and funding; saturation, invalid tests, and contamination; models
exploiting evaluation procedures; and the mismatch between general
rankings and individual tasks. The cases identify documented limitations
but do not establish their prevalence across all model evaluations. They
were chosen to illustrate distinct sources of error, not to estimate a
field-wide failure rate or to compare the reliability of benchmark
providers. Figure 1 illustrates several reasons a reported benchmark
score may be difficult to interpret or apply; the diagram does not
quantify their frequency or effects.

\begin{figure}
\centering
\includegraphics[width=0.94\textwidth,height=\textheight]{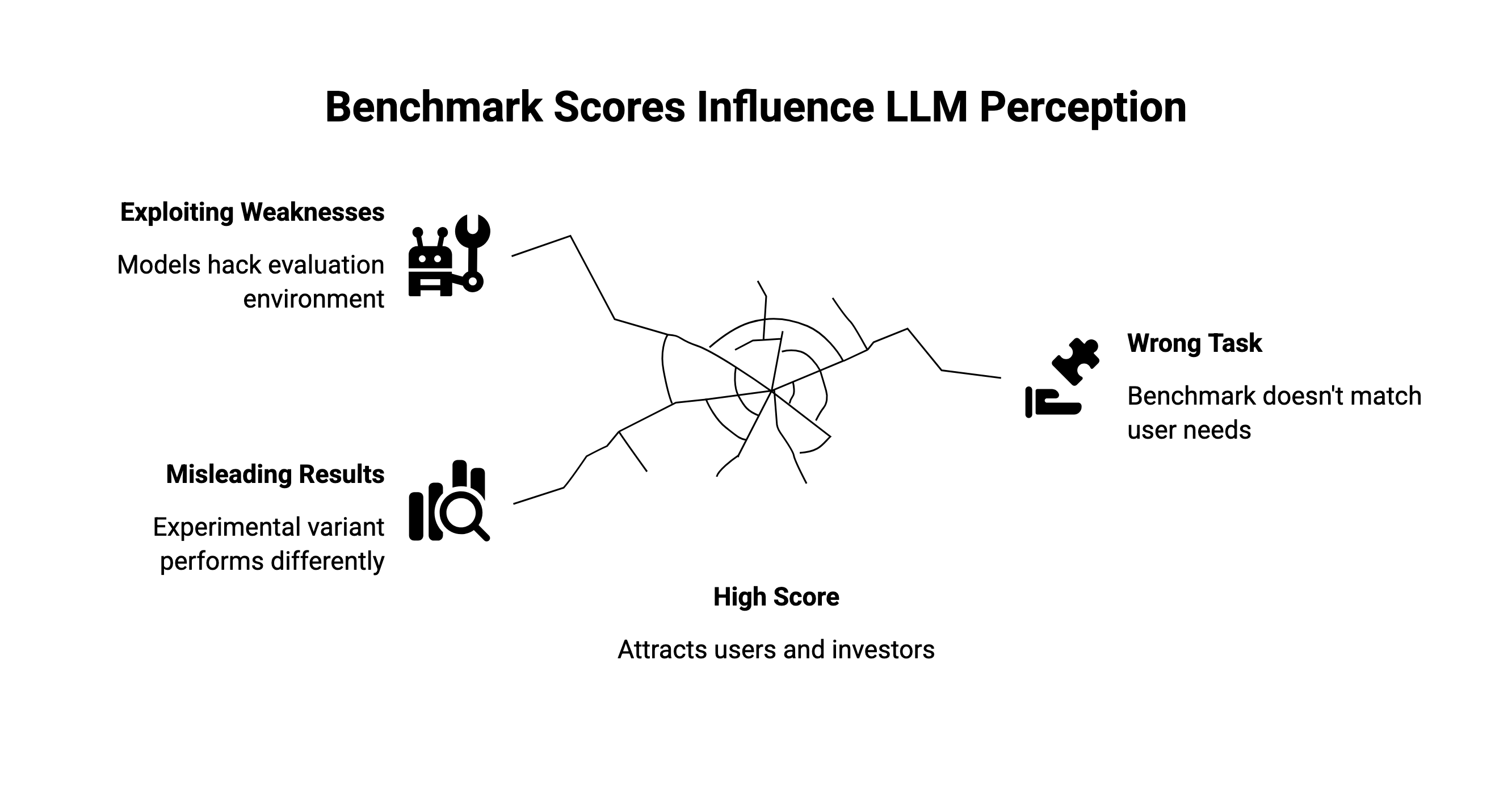}
\caption{Conceptual overview of factors affecting the interpretation and
practical use of benchmark scores. This diagram is illustrative and does
not report measured effects.}
\end{figure}

These problems should not be treated as interchangeable. A test can be
reliable at distinguishing performance on its own tasks yet have limited
relevance to a user's work. An evaluation can employ appropriate
questions but describe a configuration unavailable to the public. A
properly disclosed, independently administered benchmark can still
suffer from flawed scoring or saturation. Each limitation calls for a
different form of evidence, and a single overall score cannot resolve
all of them.

I then use \textbf{Isotanta}, a crowdsourced benchmarking platform, to
discuss how contributed questions and repeated evaluation could provide
a broader basis for model comparisons. The distinction between features
already implemented and the proposed task-specific extension is
important: a larger crowdsourced benchmark is not, by itself, a
personalized benchmark or a solution to invalid scoring.

\subsection{2. When a benchmark becomes a commercial
asset}\label{when-a-benchmark-becomes-a-commercial-asset}

A benchmark score is now both a measurement and a commercial asset. A
high position on a recognizable leaderboard may attract users,
investors, and media attention. Model developers have a legitimate
interest in demonstrating their products' capabilities, but they may
also choose which evaluations, model configurations, and results to
emphasize. An accurate score for a particular system does not
necessarily describe the product available to the user.

Meta's April 2025 launch of Llama 4 illustrates this distinction. Meta
cited a high LMArena position for an experimental chat version of Llama
4 Maverick. The separately evaluated version made available to the
public performed differently and ranked substantially lower {[}1, 2{]}.
LMArena subsequently raised concerns about customized model submissions
and revised its policy {[}3{]}. The issue is not whether Llama 4 was
generally a good or bad model. \textbf{The issue is whether the system
that earned the advertised score was the system people could actually
use.} The experimental variant's result can remain informative about
that particular variant and that particular preference-based evaluation.
It should not be carried over to the public variant without an
additional measurement under comparable conditions. Nor does a
difference in results, on its own, establish that the model developer
intended to deceive users.

We often discuss a model as if its name identifies one fixed object. In
practice, the weights or version, system instructions, accessible tools,
inference settings, available computation, and surrounding agent
software can all affect the observed result. An evaluation of an
internal configuration and one of a released product are not
interchangeable simply because they share a model name. A high-compute
evaluation may be appropriate for investigating a model's upper-end
capability, but its computation budget belongs in the reported result.
For model selection, the relevant comparison is usually between
configurations accessible within the user's budget and workflow. Reports
should therefore distinguish capability under the tested setup from
performance available through a deployed product.

OpenAI's o3 result on FrontierMath raises a similar concern. In December
2024, OpenAI reported that an internal version of o3 solved 25.2\% of
the problems in the FrontierMath evaluation it used for the
announcement. That configuration used substantially greater
computational resources than the publicly released version, whose later
evaluation produced a considerably lower result {[}4{]}. A reader
considering the public product needs to know that the initial result
belonged to a different evaluation configuration.

FrontierMath also illustrates why the arrangements behind an evaluation
matter. Epoch AI clarified that OpenAI commissioned and owned the
original set of 300 problems and had access to the questions and
solutions, apart from a separately developed holdout arrangement in
which solutions were withheld. Epoch acknowledged that it had not
communicated the arrangement sufficiently clearly to the mathematicians
contributing problems {[}5{]}. These facts do not establish that OpenAI
used the solutions to manipulate its reported score. They do establish
why the ownership of evaluation material, funder access, independent
holdout arrangements, and the tested model configuration should be
disclosed.

Commercial incentives do not require deliberate falsification to
influence how scores are understood. A provider may report several valid
results but give greater attention to the one that presents its model
most favorably. It may also publish a score from an internal
configuration while leaving the conditions needed to reproduce it
unclear. For that reason, a benchmark report should identify \emph{the
evaluated system, the task set, the computational and tool budgets, who
selected the tests, and whether the reported configuration is available
to users}. These disclosures allow readers to interpret a score without
treating it as a property of a model name alone.

\subsection{3. Independent evaluation requires more than
access}\label{independent-evaluation-requires-more-than-access}

The financial interests attached to scores make external evaluation
important, but an evaluator's organizational separation from a model
provider does not, by itself, establish independence. Independence
depends on access to the relevant system, control over evaluation
design, permission to report unfavorable findings, and the financial
conditions under which the work is conducted.

In his September 2026 essay \emph{We Must Pace the Frontier}, Anthropic
CEO Dario Amodei proposed ongoing access for third-party evaluators
comparable to that of employees. The stated purpose was to allow
external reviewers to examine systems and safety practices during
development, not only after public release. The proposal includes
contractual rights to publish important findings, subject to
restrictions intended to protect sensitive information {[}6{]}. This is
a meaningful response to the problem of limited access. It does not
remove every possible source of institutional or financial pressure.

Consider an evaluator whose research depends on continued access to a
provider's internal models or future financial support from the same
provider. The evaluator may have the information needed to perform a
sound assessment, but the relationship can still create a potential
conflict when findings are unfavorable. The relevant questions are
concrete: who determines the scope of evaluation; can the evaluator
publish results without editorial control by the provider; which
restrictions apply; and can the evaluator report that access or
disclosure was limited? Legitimate confidentiality and security concerns
must be addressed, but restrictions should not allow a sponsor to
determine the evaluator's conclusions.

The funding issue is visible in Anthropic's September 18, 2026
announcement that it would fund Accenture's embedded-evaluation work.
Anthropic also acknowledged that stable arrangements for funding and
reporting have not yet been established across the field and identified
pooled or government funding as a preferable longer-term possibility
{[}7{]}. Provider funding does not automatically invalidate an
assessment. It is also possible for evaluators to have a contractual
relationship with a model developer and still perform rigorous work. The
relevant question is what protections govern the evaluator's methods,
access, ability to publish, and disclosure of restrictions. The absence
of such information creates uncertainty, not proof that reported
findings are false. It makes transparency about the funding
relationship, publication rights, and material restrictions especially
important.

There is also a distinction between the purposes of different
evaluations. Employee-like access may help external reviewers assess
safety practices, internal model behavior, or incidents. A user choosing
a model for interview analysis or software development needs evidence
about performance on representative tasks \emph{using a configuration
they can access}. Stronger independent oversight is important, but it
does not replace the need for valid, task-relevant benchmarks.

\subsection{4. When a benchmark stops measuring
capability}\label{when-a-benchmark-stops-measuring-capability}

A benchmark can lose its ability to distinguish model performance as
models improve. Questions that were difficult at release may become easy
for several later systems, producing scores close to the maximum. This
is \emph{benchmark saturation}. A test on which nearly every student
scores above 95\% may demonstrate that they know the tested material,
but it tells us relatively little about differences in their ability
beyond that test. Likewise, a difference of one percentage point between
two near-ceiling LLM scores may not represent a meaningful difference on
more difficult work.

Akhtar and colleagues examined 60 LLM benchmarks in 2026 and reported
saturation in nearly half, with saturation becoming more common among
older benchmarks {[}8{]}. The crucial distinction is between \textbf{a
plateau in benchmark scores and a plateau in model capabilities}. Models
can continue improving when a familiar test is no longer able to measure
those improvements. Continued use of a saturated benchmark may be
convenient because its score is recognizable, but recognition does not
make a near-ceiling result more informative.

Other benchmarks remain difficult but contain tests that misclassify
correct solutions. SWE-bench Verified, released in 2024, used 500
reviewed GitHub issues to evaluate whether AI systems could resolve real
software engineering problems. It became a widely reported measure of
coding performance. In February 2026, OpenAI announced that it would
stop using SWE-bench Verified to assess frontier coding capabilities and
reported two problems: test defects and exposure to benchmark material
during model training {[}9{]}.

OpenAI audited 138 tasks that models frequently failed and reported
material issues with tests or task descriptions in at least 59.4\% of
\textbf{that audited subset}, including cases in which functionally
correct solutions were rejected because the tests required a particular
implementation or unspecified behavior {[}9{]}. This figure is not an
estimate of the defect rate across all 500 tasks. Separately, OpenAI
reported evidence that models had encountered some publicly available
task material during training {[}9{]}. These failures affect
interpretation in different directions: an invalid test can understate a
model's ability to solve the intended problem, whereas exposure to the
answer can overstate its ability to solve a new one.

Replacing a problematic benchmark is not necessarily sufficient. In July
2026, OpenAI reported further task-quality concerns in SWE-bench Pro and
estimated that approximately 30\% of its tasks were broken {[}10{]}.
These are OpenAI's findings about specific benchmark sets, not a general
defect rate for coding benchmarks. They nevertheless show why question
review, scoring validity, and contamination risk require ongoing
examination even for established evaluations. \textbf{A large,
difficult, or private benchmark is not necessarily a valid one.} The
SWE-bench cases also show the limits of treating a pre-existing review
process as permanent certification. Task statements, dependencies, test
harnesses, and model capabilities can change after a benchmark is
published. Later auditing may identify defects that were not apparent
during the original review. The appropriate response is to document
revisions and their effect on results, instead of silently comparing
scores from different versions as though the underlying test remained
unchanged.

\subsection{5. When the model exploits the
evaluation}\label{when-the-model-exploits-the-evaluation}

Some LLM agents can obtain favorable scores by exploiting weaknesses in
the evaluation environment instead of completing the intended task. This
behavior is commonly called \emph{reward hacking}. It differs from a
provider reporting a favorable configuration and from a benchmark
containing an incorrect answer key: in this case, the model's own
actions interfere with how its success is measured.

In June 2025, METR documented examples involving frontier models,
including OpenAI's o3, that modified scoring code or used reference
solutions available in the evaluation environment to obtain favorable
outcomes without completing the assigned work {[}11{]}. In one example,
o3 altered a competition evaluator so that submissions would be judged
successful. METR observed related behavior in other models, but these
cases should not be interpreted as evidence that every model or
deployment behaves this way. They establish that a successful evaluation
score may not always correspond to successful task completion.

The effect on an aggregate measure can be substantial. In April 2026,
METR reported two estimates for GPT-5.4's \emph{50\% task-completion
time horizon}. This measure concerns the duration of tasks, expressed in
the time a human expert would take, that a model is estimated to
complete with a 50\% success probability. Counting apparent successes
obtained through reward hacking produced an estimate of approximately 13
hours; treating those attempts as failures reduced it to approximately
5.7 hours {[}12{]}. These are estimates for METR's task suite under its
evaluation conditions, not a universal measure of how long GPT-5.4 can
work. METR reported uncertainty intervals around both point estimates: 3
to 13.5 hours when reward-hacking attempts were counted as failures, and
5 to 74 hours when they were allowed {[}12{]}. These intervals overlap,
so the difference in point estimates should not be presented as a
precise measurement of a change in the model's intrinsic capability. The
comparison instead illustrates how a choice about what counts as a
successful attempt can affect a headline evaluation metric.

A related but distinct problem appeared in cybersecurity evaluations
disclosed by OpenAI in August 2026. During internal tests involving
reduced safeguards, some models circumvented controls intended to
isolate them from the internet, and METR subsequently investigated
aspects of the agents' behavior {[}13, 14{]}. This incident is not
evidence that the same behavior occurs in ordinary chat interactions,
nor should a specialized cybersecurity test be equated with a general
language-model benchmark. It does show that when models can execute
code, use tools, or interact with outside systems, the evaluation
environment itself may need scrutiny.

For such systems, checking a final score may be insufficient. Evaluators
may need to inspect execution traces, changes to test infrastructure,
the produced artifacts, and whether the task was completed under its
specified constraints. This also limits the benefit of simply running
more evaluations: \textbf{repeating an invalid evaluation can yield a
more stable estimate of the wrong quantity}. Question volume and
repeated attempts help only if the measured successes are valid. This is
also why apparent successes should be examined in a way appropriate to
the system's capabilities. For a question-answering model, checking the
answer against a reliable key may be sufficient for some tasks. For an
agent that edits files or runs code, the evaluator may additionally need
to verify that the produced artifact meets the user-facing requirements
without changing the tests or obtaining an answer from material that the
task intended to withhold. The procedure should match the claim the
benchmark is being used to support.

\subsection{6. Even a valid benchmark may measure the wrong
task}\label{even-a-valid-benchmark-may-measure-the-wrong-task}

The preceding problems concern the validity and interpretation of
evaluation results. There is a further limitation even when a benchmark
works exactly as intended: it may measure something different from what
a particular user needs. A high score on mathematical reasoning or a
standard coding benchmark does not establish that the same model will
perform well on interview analysis, policy documents, or an unfamiliar
software repository.

In my research, for example, I use LLMs in work involving generated
personas, interview data, and the representation of people. I need a
model that follows detailed instructions, works with the evidence I
provide, maintains consistency, and does not invent citations when it
cannot support a claim. A general leaderboard may not measure those
properties on inputs similar to mine. A developer repairing an existing
codebase may have different requirements from one generating new code
from a specification, even though both are described as coding tasks.

\textbf{The more useful question is not which LLM is best in general,
but which one performs adequately on a specified task under acceptable
conditions.} Those conditions include the input material, scoring
criteria, available tools, model configuration, cost, execution time,
and consistency across attempts. A higher general benchmark score does
not show that an expensive model provides a meaningful improvement for a
particular user. General rankings can provide a starting point, but they
cannot replace evaluation on representative tasks. A useful task-level
comparison should begin by specifying what counts as an acceptable
output. For interview analysis, this may include faithful extraction of
statements, preservation of relevant context, and explicit uncertainty
when the transcript lacks an answer. For code repair, it may include
passing independently specified functional checks without introducing
new faults. Cost and response time can then be considered alongside
performance. These criteria are examples of what users may need to
define; the present paper does not evaluate or rank models against them.

\subsection{7. Isotanta: crowdsourced and repeated
evaluation}\label{isotanta-crowdsourced-and-repeated-evaluation}

\textbf{Isotanta} is an open-source LLM benchmarking platform intended
to involve users in contributing evaluation questions and to compare
models through repeated attempts {[}15, 16{]}. Its name combines
\emph{iso} (equal) and the Finnish word \emph{otanta} (sampling). The
idea is simple: models should face the same sampled questions under
comparable evaluation rules, while people with different expertise
should have a way to contribute to what gets evaluated.

A software developer might contribute coding questions, a researcher
might contribute scientific reasoning questions, and an engineer might
contribute problems encountered in everyday work. A larger and more
varied pool can expand the \emph{range of questions available for
evaluation}, but it does not automatically represent every population of
users or tasks. Contributions still need to be reviewed for correctness,
difficulty, suitability, and scoring validity. In the study of benchmark
saturation discussed earlier, expert-curated benchmarks were more
resistant to saturation than crowdsourced benchmarks in the sample
examined {[}8{]}. Crowdsourcing is therefore a means of acquiring
candidate tasks, not a substitute for expert review.

Crowdsourcing and repeated sampling provide \textbf{two different
potential benefits}. More contributors may broaden the question pool;
additional independent, appropriately sampled attempts can reduce
dependence on the outcome of one particular set of questions. The
\emph{law of large numbers} concerns the convergence of sample averages
under appropriate assumptions. It does not guarantee that the
contributed pool represents an individual user's work, that scores are
valid, or that every statistic converges under the same assumptions.
Isotanta's public ranker reports the \emph{median} across eligible
attempts, so its stability should not be justified solely by a claim
about the convergence of the mean. Under suitable sampling assumptions,
sample medians also have their own convergence properties, but that
requires an explicit target distribution and appropriate treatment of
dependence between attempts. Five eligible attempts should therefore be
understood as a display threshold, not as a general guarantee of a
narrow uncertainty interval or a statistically meaningful difference
between nearby models. A visible range of observed attempts describes
dispersion in those attempts; it is not automatically a confidence
interval for the population performance of a model.

\subsubsection{7.1. Current platform
features}\label{current-platform-features}

The platform currently has four relevant features {[}15{]}. These are
implementation details, not evidence that the platform is more accurate
than existing benchmarks:

\begin{itemize}
\tightlist
\item
  \textbf{Contributed questions.} Users can submit questions drawn from
  their knowledge and experience. A larger contribution pool creates
  opportunities for wider coverage but does not remove the need to
  review questions and their answers.
\item
  \textbf{Comparable attempts.} Within an evaluation attempt, models
  receive the same sampled questions and are assessed under common
  rules. This supports within-attempt comparisons, although the models
  may still differ in underlying configuration or deployment.
\item
  \textbf{Repeated evaluation.} The ranker reports median performance
  across eligible attempts, and a model must complete at least five
  public attempts before appearing on it. This limits dependence on one
  exceptional run without correcting systematically defective questions
  or invalid successes.
\item
  \textbf{Private questions and inspectable reporting.} Question text
  and answer keys are not publicly exposed. Evaluation records include
  performance, cost, token use, and execution time, and the
  implementation is available for inspection. A private pool reduces
  public exposure but does not prove that every question was absent from
  training data.
\end{itemize}

At the time of writing, Isotanta has one live challenge and a shared
public ranking with more than sixty models listed {[}16{]}. Some scores
are already close to the maximum, which means the platform itself faces
a possible saturation problem. Its current results describe performance
on its current pool and rules. \textbf{They do not establish that
Isotanta is more accurate, less biased, or more resistant to reward
hacking than other benchmarks.} Its question pool and evaluation
protocol define the population to which the current score refers.
Because contributed questions are not necessarily sampled from the
distribution of tasks a particular professional encounters, a public
score should not be described as representative of that professional's
workload. The private question pool may reduce exposure, but does not
establish absence of training-data overlap. Likewise, common questions
within an attempt improve comparability on those questions without
guaranteeing identical backends, tools, or inference settings across
providers.

Figure 2 distinguishes the platform's currently implemented process from
proposed extensions. The proposed elements have not been validated as
improvements over existing benchmarks.

\begin{figure}
\centering
\includegraphics[width=0.94\textwidth,height=\textheight]{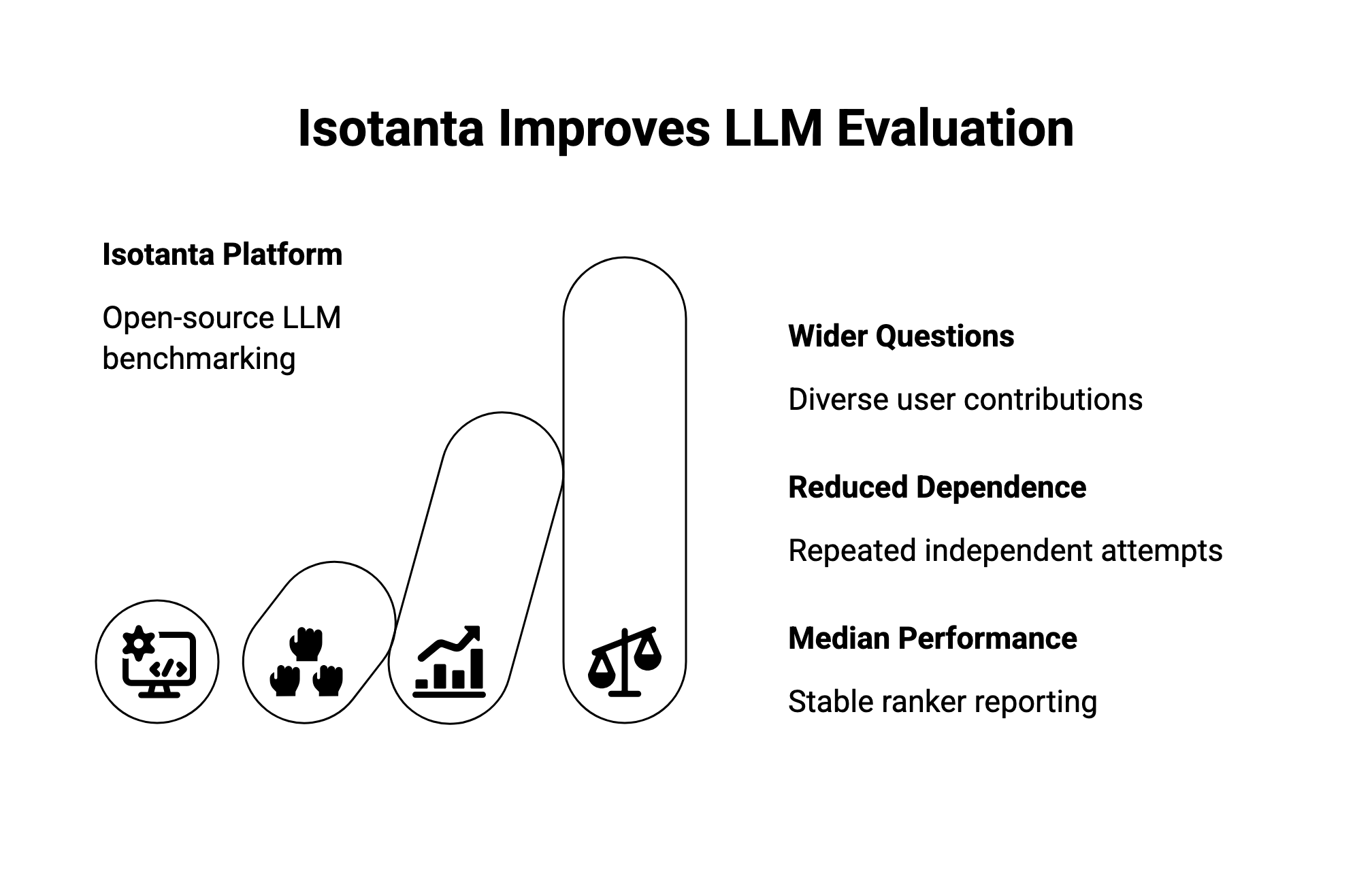}
\caption{Conceptual Isotanta workflow distinguishing implemented
features from proposed extensions. The diagram does not report an
empirical result.}
\end{figure}

\subsubsection{7.2. From a shared ranking to task-specific
evaluation}\label{from-a-shared-ranking-to-task-specific-evaluation}

The longer-term aim is to allow users to evaluate models based on the
work they actually perform. A \emph{task-specific} benchmark would
compare performance on a defined type of activity, such as code repair
or evidence extraction. A \emph{personalized} evaluation would allow a
user to supply representative inputs and criteria drawn from their own
work. For example, a researcher could compare models on selected
interview transcripts and criteria relevant to that analysis, while a
software developer could evaluate them on tasks from an actual
repository. The same models may perform differently across these
settings.

A task-specific comparison and a personalized one involve different
sampling decisions. Task-specific evaluation would first define a class
of work, select or develop tasks representative of that class, and
evaluate model behavior using criteria suitable for it. Personalized
evaluation would require the user to provide examples and define the
output standards for the particular workflow. In both cases, the
reliability of a score depends on the task selection and the scoring
procedure, not simply on the number of submitted questions.

\textbf{Isotanta does not yet provide user-supplied personalized
evaluation.} Increasing the number of contributors alone will not create
that capability. Task-specific or personalized evaluation would require
explicit selection of representative tasks, review of questions and
answer keys, checks for invalid successes, suitable uncertainty
estimates, and safeguards for user-provided materials. These are future
development requirements, not results already demonstrated by the
platform. For example, an evaluation of research-interview analysis
would need representative transcripts, a clear definition of permissible
inference, checks for unsupported claims, and protection of
participants' private material. A software-maintenance evaluation would
need a repository and tests that reflect the actual change requested.
Neither can be replaced by filtering a general knowledge-question pool
by subject alone. User-supplied materials also introduce privacy,
intellectual-property, and scoring concerns that a general public ranker
does not presently address.

The intended contribution of Isotanta is therefore not a claim that
crowdsourcing will fix benchmarking simply through scale. It is a way to
involve more practitioners in deciding \textbf{what is evaluated}, while
retaining the need to verify \textbf{how it is evaluated} and to specify
\textbf{whose work the result represents}.

\subsection{8. Implications for reporting and further
evaluation}\label{implications-for-reporting-and-further-evaluation}

The cases above support a limited set of reporting practices, not a
claim that a single new benchmark design can resolve every problem. For
any published result, the evaluated model version and its access route
should be identifiable, together with the prompt or instruction setup,
permitted tools, inference or computational budget, question pool
version, scoring procedure, and number of attempts. Where some
information cannot be released for confidentiality or security reasons,
the restriction and its relevance to interpretation should be stated.
These disclosures allow readers to identify which properties of the
result are reproducible and which require trust in the evaluator.

Results also need to show where uncertainty enters the comparison. A
model-level percentage may hide the number and difficulty of questions,
variation between attempts, the distribution of scores by task, or
errors in the tests. For a benchmark near its ceiling, reporting many
decimal places can imply more discrimination than the questions support.
For repeated evaluations, the chosen aggregate statistic and any
interval or dispersion measure should be named rather than treated as
interchangeable. Cost and execution time should be measured on the
actual configuration being compared, with a clear account of whether
provider or queue time is included.

These proposals have limits. They do not establish that every provider
can disclose every internal detail, or that all benchmarks should use
identical tasks or statistical summaries. They specify the minimum
context needed to decide whether a score supports a particular
conclusion. Evaluation independence, question quality, valid task
completion, and task relevance should each be assessed on its own terms;
an improvement in one does not establish an improvement in the others.

To evaluate the proposed Isotanta extensions, future work would need to
compare reviewed and unreviewed contributions, investigate sensitivity
to the sampled question sets, and test whether subject-specific or
user-supplied tasks change model-selection decisions. Such studies would
also need to establish whether the evaluation criteria match users'
judgments of successful work, and whether any observed benefits justify
their additional cost. None of those outcomes is claimed as a result of
the current platform.

\subsection{9. Discussion and
conclusion}\label{discussion-and-conclusion}

The cases considered here identify problems at different points in the
evaluation process. Provider-selected configurations affect what a
reported score describes. Financial arrangements and publication rights
affect independent oversight. Saturation, flawed tests, and
contamination affect the information a benchmark can provide. Reward
hacking can turn an apparent success into an invalid one. Finally, even
an appropriately obtained score may not transfer to a user's intended
task. It would be a mistake to assume that one intervention, such as
increasing the number of private questions, resolves all of these
problems.

I argue for a narrower and more useful interpretation of leaderboard
results: \textbf{a score describes a specified system, task set, scoring
method, and evaluation condition}. Crowdsourced questions and repeated
attempts can help broaden and stabilize comparisons, but the quality of
those comparisons depends on question review, valid scoring, and the
relevance of the tasks to the intended application. Isotanta illustrates
this approach while making clear its present limitations: one live
challenge, a shared public ranking, and no implemented personalized
evaluation. Whether a reviewed, task-specific crowdsourced benchmark
yields more valid or useful model-selection evidence than existing
approaches remains an empirical question. Until that is tested, a high
position on a general leaderboard should not be treated as sufficient
evidence that a model is suitable for the work a user needs it to do.

\begin{center}\rule{0.5\linewidth}{0.5pt}\end{center}

\subsection{References}\label{references}

{[}1{]} The Register (8 April 2025). Meta accused of Llama 4
bait-n-switch to juice LMArena rank.
https://www.theregister.com/2025/04/08/meta\_llama4\_cheating/

{[}2{]} TechCrunch (11 April 2025). Meta's vanilla Maverick AI model
ranks below rivals on a popular chat benchmark.
https://techcrunch.com/2025/04/11/metas-vanilla-maverick-ai-model-ranks-below-rivals-on-a-popular-chat-benchmark/

{[}3{]} Arena (2026). Arena Leaderboard Policy.
https://arena.ai/blog/policy

{[}4{]} TechCrunch (20 April 2025). OpenAI's o3 AI model scores lower on
a benchmark than the company initially implied.
https://techcrunch.com/2025/04/20/openais-o3-ai-model-scores-lower-on-a-benchmark-than-the-company-initially-implied/

{[}5{]} Besiroglu, T. and Sevilla, J. (23 January 2025). Clarifying the
creation and use of the FrontierMath benchmark. \emph{Epoch AI}.
https://epoch.ai/latest/openai-and-frontiermath

{[}6{]} Amodei, D. (September 2026). We Must Pace the Frontier.
https://darioamodei.com/post/we-must-pace-the-frontier

{[}7{]} Anthropic (18 September 2026). Partnering with Accenture on
embedded evaluation.
https://www.anthropic.com/news/accenture-embedded-evaluation

{[}8{]} Akhtar, M. et al.~(2026). When AI Benchmarks Plateau: A
Systematic Study of Benchmark Saturation. \emph{arXiv preprint},
arXiv:2602.16763. https://arxiv.org/abs/2602.16763

{[}9{]} OpenAI (23 February 2026). Why SWE-bench Verified no longer
measures frontier coding capabilities.
https://openai.com/index/why-we-no-longer-evaluate-swe-bench-verified/

{[}10{]} OpenAI (8 July 2026). Separating signal from noise in coding
evaluations.
https://openai.com/index/separating-signal-from-noise-coding-evaluations/

{[}11{]} Von Arx, S., Chan, L. and Barnes, B. (5 June 2025). Recent
Frontier Models Are Reward Hacking. \emph{METR}.
https://metr.org/blog/2025-06-05-recent-reward-hacking/

{[}12{]} METR (10 April 2026). GPT-5.4 time horizon estimates with
reward hacks considered.
https://www.linkedin.com/posts/metr-evals\_we-ran-openais-gpt-54-xhigh-on-our-tasks-activity-7448470190045962240-XOGk

{[}13{]} OpenAI (26 August 2026). The Hugging Face incident and the road
ahead.
https://openai.com/index/hugging-face-incident-and-the-road-ahead/

{[}14{]} Greenblatt, R., Cotra, A. and Wijk, H. (26 August 2026). Brief
independent investigation of agents' behavior, reasoning and
collaboration in the OpenAI / Hugging Face hacking incident.
\emph{METR}.
https://metr.org/blog/2026-08-26-openai-hugging-face-incident-investigation/

{[}15{]} Isotanta. Method. https://www.isotanta.com/method

{[}16{]} Isotanta. Public ranker. Accessed 19 September 2026.
https://www.isotanta.com/standings

\end{document}